\documentclass[letterpaper, 10 pt, journal, twoside]{IEEEtran} 

\IEEEoverridecommandlockouts

\pdfoutput=1

\usepackage[noadjust]{cite}

\usepackage{xparse} 
\usepackage{amsmath} 
\usepackage{amssymb}
\usepackage{amsfonts} 
\usepackage{xfrac}

\usepackage{graphicx}
\usepackage{subcaption}
\usepackage[keeplastbox]{flushend}
\usepackage{balance}

\makeatletter
\let\NAT@parse\undefined
\makeatother
\usepackage[colorlinks]{hyperref}
\usepackage{cleveref}

\usepackage{url}

\usepackage{booktabs}
\usepackage[para]{threeparttable}

\usepackage{xparse}

\NewDocumentCommand\bbm{}{ \begin{bmatrix} }
\NewDocumentCommand\ebm{}{ \end{bmatrix} }
\NewDocumentCommand\Vector{m}{ \boldsymbol{\mathbf{#1}} }
\NewDocumentCommand\Matrix{m}{ \boldsymbol{\mathbf{#1}} }

\NewDocumentCommand\Norm{m}{\left\Vert#1\right\Vert }

\NewDocumentCommand\PartialDerivative{mm}{ \frac{\partial #1}{\partial #2} }

\NewDocumentCommand\ArgMin{m}{ \operatorname*{argmin}_{#1} }

\NewDocumentCommand\LieGroupSE{m}{ \mathrm{SE}(#1) }

\NewDocumentCommand\Transform{}{ \Matrix{T} }

\NewDocumentCommand\Estimate{m}{\hat{#1}}

\NewDocumentCommand\Warped{m}{\tilde{#1}}

\NewDocumentCommand\Tracking{}{t}
\NewDocumentCommand\Reference{}{r}

\NewDocumentCommand\PoseReferenceToTracking{}{\Transform_{\Tracking,\Reference}}
\NewDocumentCommand\EstimatedPoseReferenceToTracking{}{\Estimate{\Transform}_{\Tracking,\Reference}}

\NewDocumentCommand\Pixel{}{\Vector{u}}
\NewDocumentCommand\ReferencePixel{}{\Pixel_{\Reference}}

\NewDocumentCommand\WarpedTrackingPixel{}{\Warped{\Pixel}_{\Tracking}}

\NewDocumentCommand\PointComponent{}{p}
\NewDocumentCommand\Point{}{\Vector{\PointComponent}}

\NewDocumentCommand\Image{}{\Vector{I}}
\NewDocumentCommand\ReferenceImage{}{\Image_{\Reference}}
\NewDocumentCommand\TrackingImage{}{\Image_{\Tracking}}
\NewDocumentCommand\ReferenceImageIndexed{m}{\Image_{\Reference}^{#1}}
\NewDocumentCommand\TrackingImageIndexed{m}{\Image_{\Tracking}^{#1}}
\NewDocumentCommand\WarpedImage{}{\Warped{\Image}}
\NewDocumentCommand\WarpedReferenceImage{}{\WarpedImage_{\Reference}}

\NewDocumentCommand\Depthmap{}{\Matrix{D}}
\NewDocumentCommand\ReferenceDepthmap{}{\Depthmap_{\Reference}}

\NewDocumentCommand\Noise{}{\Vector{n}}
\NewDocumentCommand\ImageNoise{}{\Noise_{\Image}}
\NewDocumentCommand\ImageNoiseCovariance{}{\Matrix{R}_{\Image}}
\NewDocumentCommand\DepthNoiseCovariance{}{\Matrix{R}_{\Depthmap}}
\NewDocumentCommand\ImageErrorCovariance{}{\Matrix{R}_{\ImageError}}
\NewDocumentCommand\ImageErrorJacobianDepth{}{\Matrix{G}_{\Depthmap}}

\NewDocumentCommand\ReferenceImageTransform{}{\Vector{f}}
\NewDocumentCommand\WarpedReferenceImageTransform{}{\Vector{g}}

\NewDocumentCommand\Error{}{\Vector{e}}
\NewDocumentCommand\ImageError{}{\Error_{\Image}}

\NewDocumentCommand\CameraModel{}{\Vector{\pi}}
\NewDocumentCommand\InvCameraModel{}{\CameraModel^{-1}}

\NewDocumentCommand\CNNLoss{}{\mathcal{L}}

\title{How to Train a CAT: Learning Canonical Appearance Transformations for Direct Visual Localization Under Illumination Change}

\author{Lee Clement$^{1}$ and Jonathan Kelly$^{1}$%
\thanks{Manuscript received: September, 9, 2017; Revised December, 12, 2017; Accepted January, 15, 2018.}
\thanks{This paper was recommended for publication by Editor Jana Kosecka upon evaluation of the Associate Editor and Reviewers' comments.} 
\thanks{$^{1}$Both authors are with the Space \& Terrestrial Autonomous Robotic Systems (STARS) laboratory at the University of Toronto Institute for Aerospace Studies (UTIAS), Canada. {\tt <firstname>.<lastname>@robotics.utias.utoronto.ca}}%
\thanks{Digital Object Identifier (DOI): see top of this page.}
}

\begin{document} 

\maketitle 


\markboth{IEEE Robotics and Automation Letters. Preprint Version. Accepted January, 2018}
{Clement \MakeLowercase{\textit{et al.}}: Learning Canonical Appearance Transformations for Direct Visual Localization Under Illumination Change}  

\begin{abstract}
Direct visual localization has recently enjoyed a resurgence in popularity with the increasing availability of cheap mobile computing power.
The competitive accuracy and robustness of these algorithms compared to state-of-the-art feature-based methods, as well as their natural ability to yield dense maps, makes them an appealing choice for a variety of mobile robotics applications.
However, direct methods remain brittle in the face of appearance change due to their underlying assumption of photometric consistency, which is commonly violated in practice.
In this paper, we propose to mitigate this problem by training deep convolutional encoder-decoder models to transform images of a scene such that they correspond to a previously-seen canonical appearance.
We validate our method in multiple environments and illumination conditions using high-fidelity synthetic \mbox{RGB-D} datasets, and integrate the trained models into a direct visual localization pipeline, yielding improvements in visual odometry (VO) accuracy through time-varying illumination conditions, as well as improved metric relocalization performance under illumination change, where conventional methods normally fail.
We further provide a preliminary investigation of transfer learning from synthetic to real environments in a localization context.
\end{abstract}


\section{Introduction}
\IEEEPARstart{S}{elf-localization} has long been recognized as an essential competency for autonomous vehicles, with visual localization in particular garnering significant attention over the past few decades.
Recently, direct visual localization algorithms (e.g.,~\cite{Newcombe2011-ft,Engel2015-il,Omari2015-ef,Whelan2016-jy}), which compare pixel intensities directly rather than tracking and matching abstracted features, have become popular due to their competitive accuracy to state-of-the-art indirect (feature-based) methods\footnote[2]{See, e.g., the KITTI odometry leaderboard: \url{http://www.cvlibs.net/datasets/kitti/eval_odometry.php}.}, their robustness to effects such as motion blur and camera defocus~\cite{Newcombe2011-ft}, and their natural ability to yield dense maps of an environment, which may be useful for higher-level tasks.

Despite their successes, direct methods have been hindered by their underlying assumption of photometric consistency, that is, the assumption that the observed brightness or color of objects in a scene remains constant through space and time.
In practice, this assumption makes direct localization brittle in environments containing time-varying illumination (e.g., changing shadows) or non-Lambertian materials (e.g., reflective surfaces), as well as in situations where camera parameters such as exposure and white balance may vary automatically in response to local scene brightness.
This is especially problematic when localizing under significant appearance change in long-term autonomy applications.

While some have attempted to circumvent this difficulty by treating visual localization as an end-to-end learning problem (e.g.,~\cite{Kendall2015-kh,Costante2016-yx}), such end-to-end methods have yet to prove as accurate or robust as state-of-the-art methods based on well established geometric and probabilistic modelling~\cite{Cadena2016-ds}.
On the other hand, analytical models of appearance must often make approximations or assumptions that are frequently violated in practice (e.g., photometric consistency), or require detailed knowledge of the geometry, illumination, and material properties of the environment~\cite{Whelan2016-jy,Kasper:2016vn}.

\begin{figure}
    \centering
    \includegraphics[width=0.9\columnwidth]{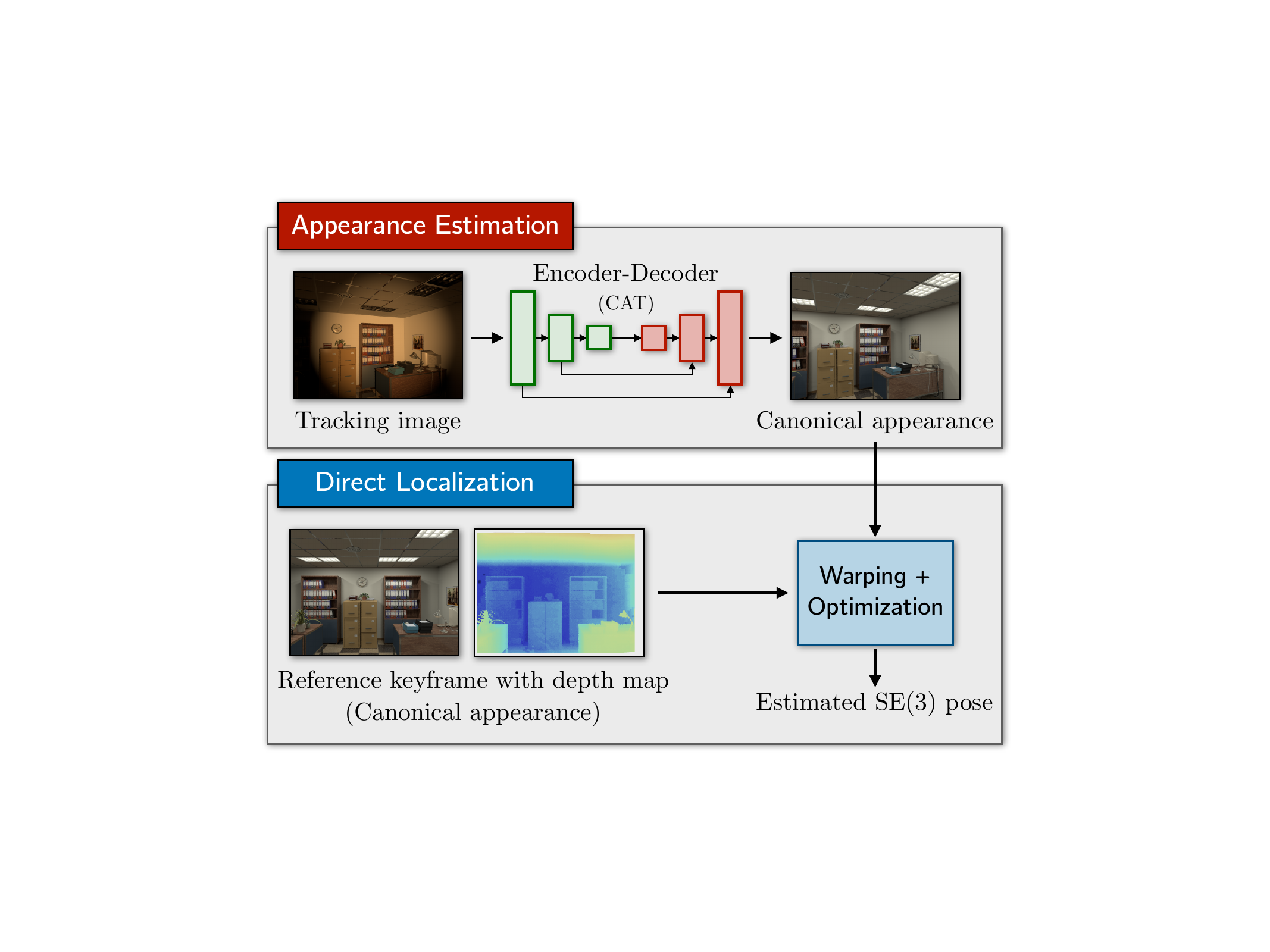}
    \caption{We train a deep convolutional encoder-decoder network to estimate the canonical appearance of a scene given an image captured under different illumination conditions, and use the transformed imagery in a direct visual localization pipeline to estimate the 6-DOF motion of the camera under illumination change.}
    \label{fig:pipeline}
    \vspace{-12pt}
\end{figure}

In this work we propose a hybrid solution for direct localization under varying illumination conditions that combines a frame-to-keyframe localization pipeline with a learned image transformation that corrects for unmodelled effects such as illumination change (\Cref{fig:pipeline}).
Rather than modelling illumination directly, we leverage existing sources of image data and recent work on image-to-image translation~\cite{Isola2016-jl} to train a deep convolutional encoder-decoder network~\cite{Hinton2006-ko,Ronneberger2015-mq} that learns to transform images of a scene that has undergone illumination change such that they correspond to a \emph{canonical appearance} of the scene (i.e., a previously-seen reference condition).
We refer to this learned transformation as a \emph{canonical appearance transformation} (CAT).
Using high-fidelity synthetic \mbox{RGB-D} datasets, we demonstrate that our method yields significant improvements in visual odometry (VO) accuracy under time-varying illumination, as well as improved tracking performance in keyframe-based relocalization under conditions of severe illumination change, where conventional methods often fail.
We further provide a preliminary investigation of transfer learning from synthetic to real environments in a localization context.
An open-source implementation of our method using PyTorch is available at \url{https://github.com/utiasSTARS/cat-net}.

\section{Related Work}
Illumniation robustness in visual localization has been previously studied from the perspective of illumination invariance, with methods such as~\cite{McManus2014-op,Paton2017-fi,Clement2017-gx} making use of hand-crafted image transformations to improve feature matching over time.
Similarly, affine models~\cite{Engel2015-il} and other analytical transformations~\cite{Park2017-zx} have been used to improve the robustness of direct visual localization to illumination change.
However, there has been little work on using machine learning techniques to generate such models from data.

The use of machine learning as a complement to analytically derived models has met with considerable success in the field of optimal control (e.g.,~\cite{Levine2016-end,2017_Li_Deep}), and has recently gained a foothold in the domain of state estimation.
Techniques such as~\cite{Kendall2015-kh,Costante2016-yx} learn end-to-end estimators using deep learning, while others such as~\cite{Handa2016-tn,Haarnoja2016-mg} combine deep models with traditional estimation machinery.
Still other work has focused on training deep models to extract illumination information from images, which can be used to improve the performance of traditional localization systems~\cite{Peretroukhin2017-lt,Ma2017-gr}.  

Our work is related to the field of image-based rendering (e.g.,~\cite{Fitzgibbon2005-bu,Flynn2016-fy}), which aims to synthesize new views of a scene by blending existing images.
In particular, our work bears resemblances to~\cite{Flynn2016-fy}, which generates such imagery using a convolutional neural network (CNN).
In contrast to~\cite{Flynn2016-fy}, our goal is not to learn an image synthesis pipeline to be queried at arbitrary poses, but rather to learn a correction to the appearance of a single image taken from a fixed pose.

The closest work to ours is~\cite{Gomez-Ojeda2017-ai}, which proposes to use deep models to enhance the temporal consistency and gradient information of image streams captured in environments with high dynamic range.
In such environments, the main source of appearance change is the camera itself as it automatically modulates its imaging parameters in response to the local brightness of a static environment.
In contrast, our method is concerned with improving localization under \emph{environmental} illumination change, and is equally applicable to visual odometry (VO) and visual relocalization tasks.

\section{Direct Visual Localization}
We adopt a keyframe-based direct visual localization pipeline similar to~\cite{Omari2015-ef}, which is locally drift-free and suitable for use with stereo or \mbox{RGB-D} cameras.
Our method uses an inverse compositional warping function to implicitly establish correspondences between a reference image (keyframe) and a tracking (live) image, and minimizes the photometric error between corresponding pixels to estimate the relative pose of the camera.
Keyframes are created whenever the translational or rotational distance between the tracking image and the active keyframe exceed a preset threshold.

\subsection{Observation model} \label{sec:localization:obsmodel}
Assuming that our images have been undistorted and rectified through an appropriate calibration, we can approximate our camera by a pinhole model with focal lengths $f_u, f_v$ and principal point $(c_u, c_v)$.
Thus, our (noiseless) observation model mapping 3D point $\Point = \bbm \PointComponent_x & \PointComponent_y & \PointComponent_z \ebm^T$ onto image coordinates $\Pixel = \bbm u & v \ebm^T$ and depth map $\Depthmap$ is given by
\begin{align} \label{eq:camera_model}
    \bbm \Pixel \\ \Depthmap(\Pixel) \ebm &= \CameraModel(\Point) 
            = \bbm f_u \PointComponent_x / \PointComponent_z + c_u \\ f_v \PointComponent_y / \PointComponent_z + c_v  \\ \PointComponent_z \ebm,
\end{align}
and the inverse mapping is given by
\begin{align} \label{eq:inv_camera_model}
    \Point &= \InvCameraModel( \Pixel, ~\Depthmap(\Pixel) )
                    = \Depthmap(\Pixel) \bbm (u-c_u) / f_u \\ (v-c_v)  / f_v \\ 1 \ebm.
\end{align}
Note that $\Depthmap$ is easily replaced by a disparity map in the case of stereo vision, with only minor modifications to the model.

Using \Cref{eq:camera_model,eq:inv_camera_model}, we can then map the image coordinates $\ReferencePixel$ of a reference image $\ReferenceImage$ onto the warped image coordinates $\WarpedTrackingPixel$ of a tracking image $\TrackingImage$ given the reference depth map $\ReferenceDepthmap$:
\begin{align}
    \WarpedTrackingPixel &= \CameraModel \left( \PoseReferenceToTracking ~\InvCameraModel\left( \ReferencePixel, ~\ReferenceDepthmap(\ReferencePixel) \right) \right)
\end{align}
where $\PoseReferenceToTracking \in \LieGroupSE{3}$ is the 6-DOF pose of the tracking image relative to the reference image (i.e., the pose we want to estimate).
Note that through a slight abuse of notation we treat the Cartesian and homogeneous coordinates of $\Point$ interchangeably.
Finally, we can compute a reconstruction $\WarpedReferenceImage$ of the reference image by sampling the tracking image:
\begin{align} \label{eq:sampling}
    \WarpedReferenceImage(\ReferencePixel) &= \TrackingImage(\WarpedTrackingPixel).
\end{align}


\subsection{Photometric (in)consistency}
Direct visual localization typically relies on the assumption of \emph{photometric consistency} to compute the error terms to be minimized.
In other words, we assume that the observed brightness or color of the scene stays constant across the reference and tracking images.
In our inverse compositional formulation, this assumption can be expressed as
\begin{align} \label{eq:photometric_consistency}
    \ReferenceImage(\Pixel) = \WarpedReferenceImage(\Pixel) + \ImageNoise,
\end{align}
where $\WarpedReferenceImage$ is the reconstruction of reference image $\ReferenceImage$ based on the current pose estimate $\PoseReferenceToTracking$, and $\ImageNoise$ is zero-mean Gaussian noise with covariance $\ImageNoiseCovariance$.

In practice, the observed brightness of a scene may vary for a variety of reasons.
For example, modern cameras will automatically adjust their gain and white balance parameters in response to the local brightness of the scene.
While it is possible to modulate these parameters~\cite{Zhang2017-sv} or estimate their effect using an affine model~\cite{Engel2015-il,Park2017-zx} or calibrated camera response function~\cite{Engel2017-tq}, the response of the camera is not the only source of variation.
Crucially, the illumination of an otherwise static scene may vary with time, or the materials in the scene may exhibit non-Lambertian reflectance. 
These effects are difficult to model analytically (although some attempts have been made in this context, e.g.,~\cite{Whelan2016-jy,Kasper:2016vn}) and require detailed knowledge of the geometry, illumination, and material properties of the scene.

To account for variations in observed brightness, we can generalize \Cref{eq:photometric_consistency} so that it has the form
\begin{align} \label{eq:generalized_photometric_consistency}
    (\ReferenceImageTransform \circ \ReferenceImage)(\Pixel) &= (\WarpedReferenceImageTransform \circ \WarpedReferenceImage)(\Pixel) + \ImageNoise,
\end{align}
where $\circ$ denotes function composition.
In other words, we wish to find functions $\ReferenceImageTransform(\cdot)$ and $\WarpedReferenceImageTransform(\cdot)$ that maximize the photometric consistency of the two images for a given $\PoseReferenceToTracking$.
In this work, we choose $\ReferenceImageTransform(\cdot) \equiv \WarpedReferenceImageTransform(\cdot)$, and learn an approximation of $\ReferenceImageTransform(\cdot)$ from data.
Specifically, we learn an $\ReferenceImageTransform(\cdot)$ that transforms both images such that they correspond to a chosen \emph{canonical appearance}, such as static diffuse illumination.
This formulation captures two problematic cases for direct localization: first, it provides a means of enhancing the temporal consistency of the image stream (similar to~\cite{Gomez-Ojeda2017-ai}), which can improve the accuracy and robustness of VO under time-varying illumination; and second, it allows us to create a map of an environment under nominal illumination conditions, then relocalize against it under different conditions.

\subsection{Relative motion estimation}
With $\ReferenceImageTransform(\cdot) \equiv \WarpedReferenceImageTransform(\cdot)$, we can compute the per-pixel error terms from \Cref{eq:generalized_photometric_consistency} as
\begin{align} \label{eq:image_error}
    \ImageError(\Pixel) = (\ReferenceImageTransform \circ \ReferenceImage)(\Pixel) - (\ReferenceImageTransform \circ \WarpedReferenceImage)(\Pixel)
\end{align}
and the Jacobian of $\ImageError$ with respect to $\PoseReferenceToTracking$ as
\begin{align}
    \PartialDerivative{\ImageError(\Pixel)}{\PoseReferenceToTracking} &= 
        -\PartialDerivative{(\ReferenceImageTransform \circ \WarpedReferenceImage)(\Pixel)}{\Pixel}
        \PartialDerivative{\Pixel}{\Point}
        \PartialDerivative{\Point}{\PoseReferenceToTracking}.
\end{align} 
Note that we can obtain $\PartialDerivative{(\ReferenceImageTransform \circ \WarpedReferenceImage)(\Pixel)}{\Pixel}$ directly from the transformed image rather than by differentiating $\ReferenceImageTransform(\cdot)$.

With these quantities in hand we can compute an estimate $\EstimatedPoseReferenceToTracking$ of the relative camera pose by using Gauss-Newton optimization to solve the nonlinear least squares problem,
\begin{align} \label{eq:pose_optimization}
    \EstimatedPoseReferenceToTracking &= \ArgMin{\PoseReferenceToTracking} \sum_{\Pixel} \ImageError^T \ImageErrorCovariance^{-1} \ImageError,
\end{align}
where $\ImageErrorCovariance$ is the covariance of $\ImageError$ and we have omitted the dependence on $\Pixel$ for notational convenience.
As $\ImageErrorCovariance$ encapsulates the noise properties of both the camera sensor and the depth sensor (or disparity map computation), it can vary from pixel to pixel.
Accordingly, we estimate $\ImageErrorCovariance$ as 
\begin{align}
    \ImageErrorCovariance = \ImageNoiseCovariance + \ImageErrorJacobianDepth \DepthNoiseCovariance \ImageErrorJacobianDepth^T,
\end{align}
where $\ImageNoiseCovariance$ is the covariance of $\ImageNoise$ in \Cref{eq:generalized_photometric_consistency}, $\DepthNoiseCovariance$ is the per-pixel covariance of the depth map, 
\begin{align}
    \ImageErrorJacobianDepth = \PartialDerivative{\ImageError}{\Depthmap} = -\PartialDerivative{(\ReferenceImageTransform \circ \WarpedReferenceImage)}{\Pixel}
    \PartialDerivative{\Pixel}{\Point}
    \PartialDerivative{\Point}{\Depthmap}
\end{align}
is the Jacobian of the per-pixel error with respect to the depth map, and we have again omitted the dependence on $\Pixel$.

In line with previous work~\cite{Engel2015-il}, we apply a robust Huber loss function to \Cref{eq:pose_optimization} to mitigate the effect of outliers and false correspondences, and solve the modified problem using the method of iteratively reweighted least squares.
We refer the reader to~\cite{Barfoot2017-ri} for the full solution details, including the special treatment required to appropriately handle the $\LieGroupSE{3}$ manifold on which $\PoseReferenceToTracking$ resides.

\subsection{Keyframe mapping and relocalization}
Our direct localization pipeline operates in both mapping (VO) and relocalization modes in a similar vein to topometric visual teach-and-repeat navigation~\cite{Paton2017-fi,Clement2017-gx}, where the camera follows a similar trajectory during both mapping and relocalization phases.
As the camera explores the environment in mapping mode, we generate a list of posed keyframes with corresponding image and depth data, creating new keyframes when the translational or rotational distance between the most recent keyframe pose and the current tracking pose exceeds a preset threshold.
In relocalization mode, the system is initialized with an active keyframe and an initial guess of the camera pose, and continuously identifies and localizes against the nearest keyframe (in the Euclidean sense).

\subsection{Practical considerations}
The cost function described in~\Cref{eq:pose_optimization} is highly non-convex, and special care must be taken to avoid local minima.
Like~\cite{Newcombe2011-ft,Engel2015-il,Omari2015-ef}, we adopt a coarse-to-fine optimization scheme in which we iteratively re-solve~\Cref{eq:pose_optimization} at multiple scales in a Gaussian image pyramid to improve the efficiency and convergence radius of the problem.
We traverse the pyramid from lowest to highest resolution, using the solution at each scale as the initial guess to the next.
Following the advice of~\cite{Newcombe2011-ft}, we optimize only for rotation at the lowest resolution.
Similarly to~\cite{Engel2015-il}, we consider only pixels whose gradient magnitude exceeds a certain threshold, which allows us to smoothly trade off computation time against information usage.
Finally, we make the approximation
\begin{align} \label{eq:image_jacobian_approximation}
    \PartialDerivative{(\ReferenceImageTransform \circ \WarpedReferenceImage)(\Pixel)}{\Pixel} \approx \PartialDerivative{(\ReferenceImageTransform \circ \ReferenceImage)(\Pixel)}{\Pixel},
\end{align}
under the assumption of small camera motion, which allows us to compute the image Jacobian for keyframes only~\cite{Omari2015-ef}.

\begin{figure*}
    \centering
    \includegraphics[width=0.85\textwidth]{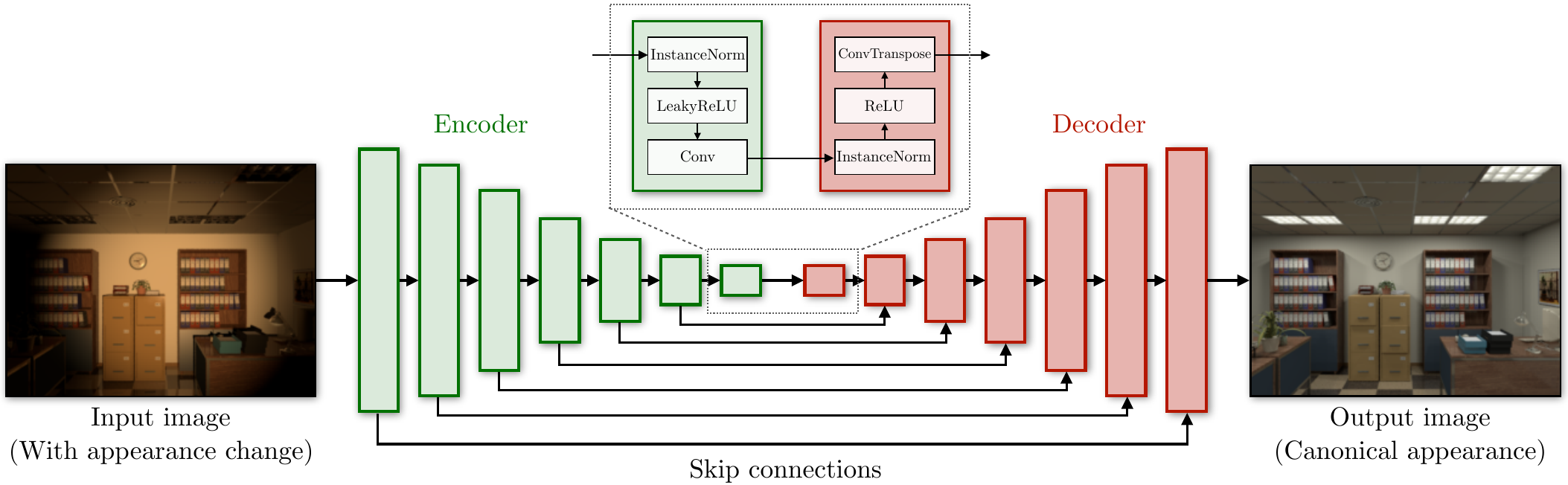}
    \caption{Our network architecture is a U-Net based on~\cite{Isola2016-jl}, consisting of seven encoder blocks (green) and seven decoder blocks (red), each sharing information with their counterpart blocks, which takes as input a $256 \times 192 \times 3$ image and outputs an image of the same dimensions. The input image is downsampled in each encoding block such that it is reduced to a $1 \times 1$ feature map at the bottleneck. Each encoder block consists of instance normalization, LeakyReLU activation, and stride-2 convolutions (downsampling), while each decoder block consists of instance normalization, ReLU activation, and stride-$\sfrac{1}{2}$ convolutions (upsampling).}
    \label{fig:unet}
    \vspace{-12pt}
\end{figure*}

\section{Learning Canonical Scene Appearance}
We now turn to the problem of finding an appropriate function $\ReferenceImageTransform(\cdot)$ that minimizes \Cref{eq:image_error} for a fixed $\PoseReferenceToTracking$.
While there are many approaches to finding $\ReferenceImageTransform(\cdot)$, an appealing choice for handling the complexity of visual data is to train an encoder-decoder model~\cite{Hinton2006-ko} based on deep convolutional neural networks (CNNs).
A typical architecture for such a model consists of a compression stage that convolves an input image with a battery of learned filters and downsamples it to a minimal representation, followed by a decompression stage that upsamples and `deconvolves' this representation to construct a new image of the original size. 

Following recent work on image-to-image translation~\cite{Isola2016-jl}, we adopt a variant of the encoder-decoder network called a \emph{U-Net}~\cite{Ronneberger2015-mq}, which is augmented with skip connections between corresponding blocks in the compression and decompression stages.
This allows the network to preserve information that may otherwise have been lost during the compression stage.

\Cref{fig:unet} shows our model architecture, consisting of seven encoder blocks (green) and seven decoder blocks (red), each sharing information with their counterparts. Each encoder block consists of instance normalization, LeakyReLU activation, and stride-2 convolutions (downsampling), while each decoder block consists of instance normalization, ReLU activation, and stride-$\sfrac{1}{2}$ convolutions (upsampling). 
The outermost blocks are exceptions: the first encoder block consists of only strided convolutions, while the final decoder block constrains the range of the output image to $[0,1]$ by applying a scaled and translated $\tanh(\cdot)$. 
We include dropout layers after the three innermost encoder and decoder blocks to reduce overfitting.

We train our model on pairs of corresponding images captured at identical poses under different illumination conditions, and apply this transformation to the incoming image stream directly.
Practically, at present, this limits us to training on synthetic datasets since, to our knowledge, no real-world dataset exists that provides calibrated stereo or \mbox{RGB-D} images captured under variable illumination at identical poses over long trajectories.
However, there is mounting evidence that models trained on synthetic data can nevertheless be useful for vision tasks in real environments~\cite{Gaidon2016-by,Peris2012-nh,Skinner2016-sh,Tobin2017-rx}.
Accordingly, we provide a preliminary investigation of synthetic-to-real transfer learning using data from real-world environments similar to the synthetic training environments.

\subsection{Datasets}
We explored our method using two synthetic \mbox{RGB-D} datasets created using high-fidelity rendering techniques:

\begin{figure*}
    \centering
    \begin{subfigure}{0.61\textwidth}
        \includegraphics[width=\textwidth]{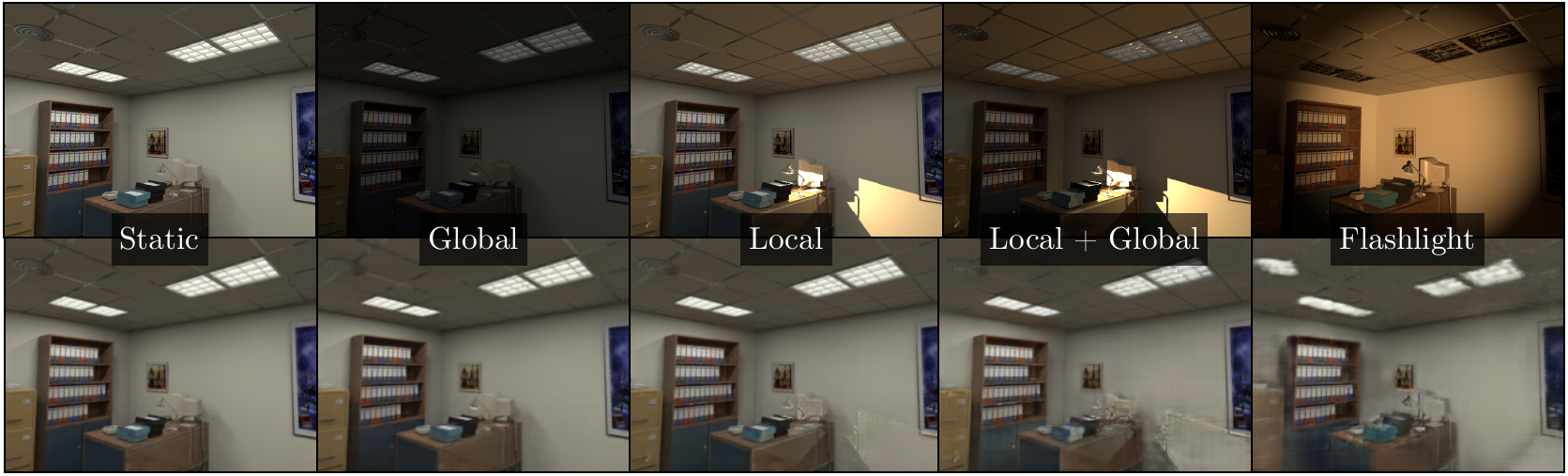}
        \caption{Synthetic sequences: \texttt{ETHL/syn1}}
        \label{fig:ethl_syn}
    \end{subfigure}
    ~
    \begin{subfigure}{0.365\textwidth}
        \includegraphics[width=\textwidth]{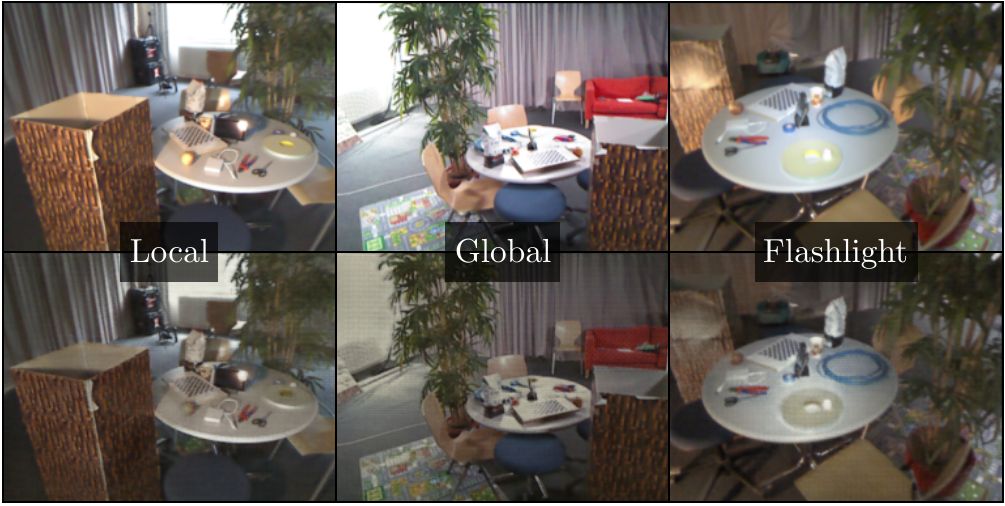}
        \caption{Real sequences: \texttt{ETHL/real}}
        \label{fig:ethl_real}
    \end{subfigure}
    \caption{\emph{Top row:} Sample images from (a) the \texttt{ETHL/syn1} sequences captured at the same pose under different illumination conditions, and (b) the \texttt{ETHL/real} sequences captured at different poses under different illumination conditions~\cite{Park2017-zx}. \emph{Bottom row:} The same images with a learned transformation that attempts to re-illuminate each image under the ``Static'' condition of the \texttt{ETHL/syn} environment.}
\end{figure*}


\begin{figure*}
    \centering
    \begin{subfigure}{0.55\textwidth}
        \includegraphics[width=\textwidth]{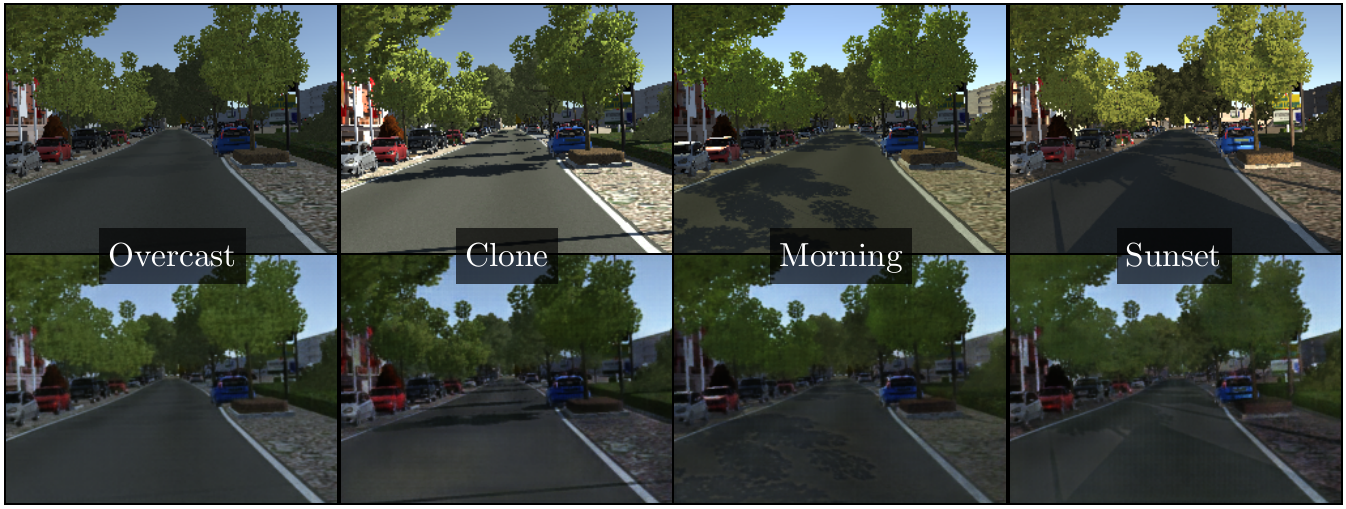}
        \caption{Synthetic sequences: \texttt{VKITTI/0001}}
        \label{fig:vkitti}
    \end{subfigure}
    ~
    \begin{subfigure}{0.412\textwidth}
        \includegraphics[width=\textwidth]{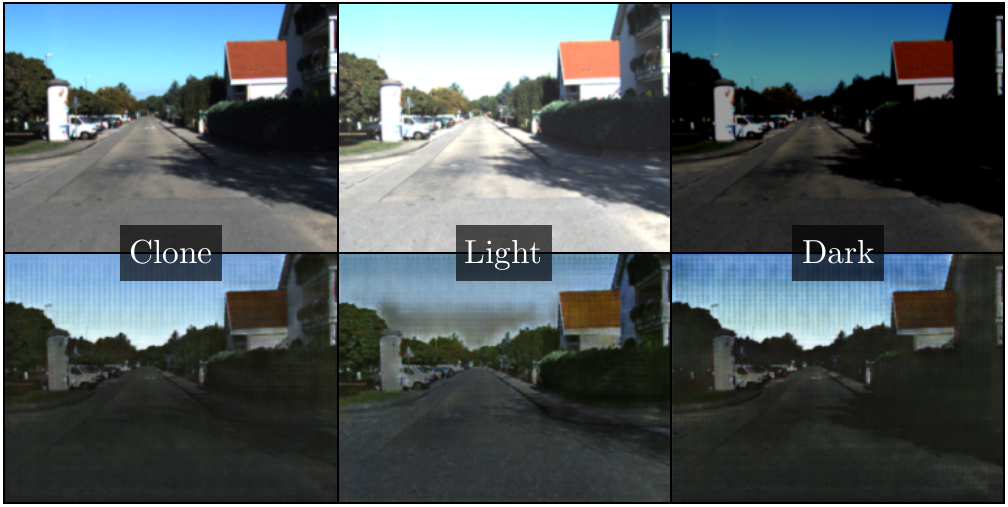}
        \caption{Real (synthetically modified) sequences: \texttt{KITTI/05}}
        \label{fig:kitti_affine}
    \end{subfigure}
    \caption{\emph{Top row:} Sample images from (a) the \texttt{VKITTI/0001} sequences and (b) the \texttt{KITTI/05} sequences captured at the same pose under different illumination conditions~\cite{Geiger2013-ky,Gaidon2016-by}. \emph{Bottom row:} The same images with a learned transformation that attempts to re-illuminate each image under the ``Overcast'' condition of the \texttt{VKITTI} environment.}
    \vspace{-12pt}
\end{figure*}


\subsubsection{ETHL Dataset}
The recent \texttt{ETHL} dataset~\cite{Park2017-zx} consists of three sets of camera trajectories and corresponding \mbox{RGB-D} imagery ($640 \times 480$ resolution) captured under various illumination conditions.
Two of these sets, dubbed \texttt{ETHL/syn1} and \texttt{ETHL/syn2}, consist of images captured along two different trajectories under five different illumination conditions, all in a simulated environment based on the ICL-NUIM dataset~\cite{Handa2014-pk}.
The ``Local'', ``Global'', and ``Local/Global'' conditions consist of time-varying illumination, while the ``Flashlight'' condition consists of view-dependent illumination generated by a light source attached to the camera (\Cref{fig:ethl_syn}).
The third set, dubbed \texttt{ETHL/real}, consists of three different camera trajectories with VICON ground truth captured in a cluttered desk scene under illumination conditions analogous to those found in the \texttt{ETHL/syn} sequences (\Cref{fig:ethl_real}).
We note that the appearance and illumination of the \texttt{ETHL/real} scene differ significantly from those found in the synthetic scenes.

\subsubsection{Virtual KITTI Dataset} \label{sec:vkitti_dataset}
The Virtual KITTI (\texttt{VKITTI}) dataset~\cite{Gaidon2016-by} is a partial virtual reconstruction of the KITTI vision benchmark~\cite{Geiger2013-ky}, consisting of five sets of camera trajectories with \mbox{RGB-D} imagery  ($1242 \times 375$ resolution) rendered under a variety of simulated illumination conditions, including ``Morning'', ``Sunset'', ``Overcast'' and ``Clone'' (\Cref{fig:vkitti}). 
The latter is meant to replicate the conditions found in the original data.
This dataset has previously been used to demonstrate transfer learning between real and synthetic environments for visual object tracking~\cite{Gaidon2016-by}.

The KITTI vision benchmark itself does not contain trajectories with significant overlap, nor does it exhibit variation in environmental illumination since each sequence was recorded around the same time of day.
We nonetheless investigated synthetic-to-real transfer learning by creating three copies of the 2.2 km KITTI odometry benchmark sequence \texttt{KITTI/05}, modified by a per-pixel affine transformation $\Image^\prime(\Pixel) = a \Image(\Pixel) + b$ (\Cref{fig:kitti_affine}). 
While an affine transformation has no impact on directional effects such as shadows or reflectance, it is analagous to a global change in illumination intensity similar to the ``Global'' condition of the \texttt{ETHL} sequences.
We refer to these conditions as ``Clone'' ($a =1, b=0$), ``Light'' ($a=1.5,b=0.1$) and ``Dark'' ($a=0.8,b=-0.2$).

\subsection{Training}
We generated training image pairs consisting of a target image captured under a chosen canonical illumination condition and a corresponding input image captured under different conditions at the same pose.
We chose the ``Static'' illumination condition as the canonical appearance of the \texttt{ETHL} sequences, and the ``Overcast'' condition as the canonical appearance of the \texttt{VKITTI} sequences.
During training, we resize and center-crop each image to $320 \times 240$ resolution, then apply a random crop to obtain image pairs of the desired $256 \times 192$ resolution.
This random cropping step provides an easy way to augment the dataset and reduce overfitting by ensuring that different data are used in each training epoch.

While~\cite{Isola2016-jl} combines an $L_1$ loss with an adversarial loss~\cite{Goodfellow2014-df} to learn a mapping between input and output images that maximizes a subjective measure of realism, in our case we are interested in learning a mapping that explicitly maximizes the photometric consistency of the output and target images in an $L_2$ sense (see~\Cref{eq:pose_optimization}).
We therefore trained the U-Net directly using the squared $L_2$ loss:
\begin{align}
    \CNNLoss = \frac{1}{N W H C} \sum_{i=1}^N \sum_{\Pixel} \Norm{ \ReferenceImageIndexed{i}(\Pixel) - (\ReferenceImageTransform \circ \TrackingImageIndexed{i})(\Pixel) }_2^2,
\end{align}
where $N=64$ is our chosen batch size, and $W$, $H$, and $C$ are the width, height, and channels of $\ReferenceImageIndexed{i}$, respectively ($256 \times 192 \times 3$).
We trained each of our models from scratch for 100 epochs, using the Adam optimizer~\cite{Kingma2015-wl} with a learning rate of $10^{-4}$ and other parameters identical to~\cite{Isola2016-jl}.

\begin{table*}[]
    \centering
    \caption{Comparison of direct visual odometry (VO) results under rapidly time-varying illumination in the \texttt{ETHL} sequences, with and without applying a learned canonical appearance transformation (CAT). The best results are highlighted in bold.}
    \label{tab:vo}
    \begin{threeparttable}
    \begin{tabular}{@{}llcccccccc@{}}
        \toprule
        &  & \multicolumn{2}{c}{\textbf{Frames Tracked (\%)}} &  & \multicolumn{2}{c}{\textbf{Avg. Trans. Err. (\% Dist.)}} &  & \multicolumn{2}{c}{\textbf{Avg. Rot. Err. ($\mathbf{\times 10^{-2}}$ deg/m)}} \\ \cmidrule{3-4} \cmidrule{6-7} \cmidrule{9-10}
       \multicolumn{2}{l}{\textbf{Sequence (length)}} & Without CAT & With CAT &  & Without CAT & With CAT &  & Without CAT & With CAT \\ \midrule
       \multicolumn{2}{l}{\texttt{ETHL/syn1} (880 frames, 9.0 m)} &  &  &  &  &  &  &  &  \\
        & Static (canonical) & 100.00 & 100.00 &  & \textbf{1.44} & 1.55 &  & \textbf{44.73} & 45.17 \\
        & Local & 100.00 & 100.00 &  & 3.22 & \textbf{1.78} &  & 115.21 & \textbf{45.06} \\
        & Global & 100.00 & 100.00 &  & 4.88 & \textbf{1.55} &  & 227.86 & \textbf{49.28} \\
        & Local + Global & 100.00 & 100.00 &  & 4.44 & \textbf{1.78} &  & 135.29 & \textbf{61.71} \\
        & Flashlight & 100.00 & 100.00 &  & 36.85 & \textbf{7.88} &  & 1092.67 & \textbf{178.25} \\ \addlinespace
       \multicolumn{2}{l}{\texttt{ETHL/syn2} (1240 frames, 7.8 m)} &  &  &  &  &  &  &  &  \\
        & Static (canonical) & 100.00 & 100.00 &  & \textbf{1.66} & \textbf{1.66} &  & 32.35 & \textbf{31.07} \\
        & Local & 100.00 & 100.00 &  & 4.22 & \textbf{1.66} &  & 111.00 & \textbf{32.74} \\
        & Global & 100.00 & 100.00 &  & 7.93 & \textbf{1.66} &  & 165.35 & \textbf{30.95} \\
        & Local + Global & 100.00 & 100.00 &  & 7.42 & \textbf{1.66} &  & 139.00 & \textbf{31.84} \\
        & Flashlight & 100.00 & 100.00 &  & 23.66 & \textbf{6.91} &  & 1293.22 & \textbf{98.85} \\ \addlinespace
       \multicolumn{2}{l}{\texttt{ETHL/real}\tnote{1}} &  &  &  &  &  &  &  &  \\
        & Local (1455 frames, 13.9 m) & 100.00 & 100.00 &  & 2.30 & \textbf{1.51} &  & 102.30 & \textbf{87.79} \\
        & Global (1396 frames, 22.1 m) & 100.00 & 100.00 &  & 3.62 & \textbf{3.48} &  & 125.75 & \textbf{121.76} \\
        & Flashlight (1387 frames, 12.1 m) & 100.00 & 100.00 &  & 3.39 & \textbf{2.81} &  & 185.63 & \textbf{182.82} \\ \bottomrule
    \end{tabular}
    \begin{tablenotes}
        \item[1] Model trained on all \texttt{ETHL/syn1} and \texttt{ETHL/syn2} sequences. Data from \texttt{ETHL/real} sequences were not used for training the models used on the \texttt{ETHL/syn} sequences.
    \end{tablenotes}
    \end{threeparttable}
    \vspace{-12pt}
\end{table*}

\subsection{Testing}
During testing, we do not use random cropping, but rather resize and center-crop each input image directly to $256 \times 192$ resolution.
\Cref{fig:ethl_syn,fig:vkitti} show sample inputs (top rows) and outputs (bottom rows) of our trained models for each illumination condition of the \texttt{ETHL/syn1} and \texttt{VKITTI/0001} sequences.
The models capture the low-frequency components of illumination change well, but struggle to render high frequency texture or completely deal with other high-frequency effects such as strong shadows.
As a result, the network outputs are often slightly blurred compared to the original images, and sometimes contain artifacts in regions of significant local appearance change.
However, the impact of such artifacts is partially mitigated by our use of a robust loss function in our localization algorithm.

\section{Visual Localization Experiments}
We conducted several visual localization experiments to validate the use of our canonical appearance transformation (CAT) models in our direct localization pipeline.
Specifically, we examined the accuracy and success rates of both visual odometry under time-varying illumination and metric localization against a keyframe map under subsequent illumination change.
In each experiment, we trained a model using all available imagery from the other trajectories in the dataset, and tested it on the remaining trajectory (e.g., the model used for the four \texttt{VKITTI/0001} sequences was trained on the remaining sixteen \texttt{VKITTI} sequences listed in \Cref{tab:localization}).
Each training set contained roughly 4000--8000 image pairs.

\subsection{Visual odometry (VO)}
To assess the usefulness of our trained models for doing VO through time-varying illumination, we compared the performance of our direct VO pipeline on each \texttt{ETHL} sequence individually, both with and without applying a CAT model.

\begin{figure}
    \centering

    \begin{subfigure}{0.9\columnwidth}
        \includegraphics[width=\textwidth]{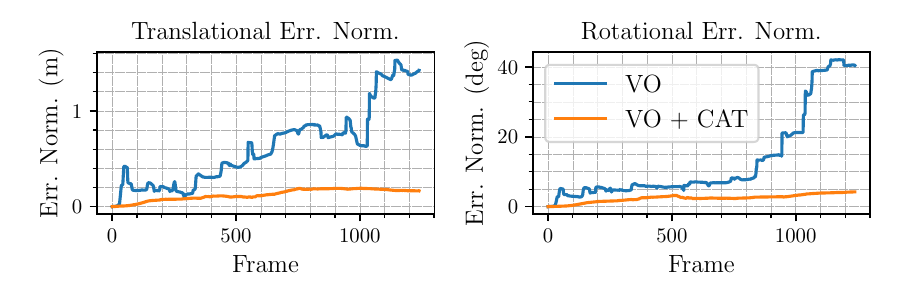}
    \end{subfigure}
    \vspace{-5pt}

    \begin{subfigure}{0.9\columnwidth}
        \includegraphics[width=\textwidth]{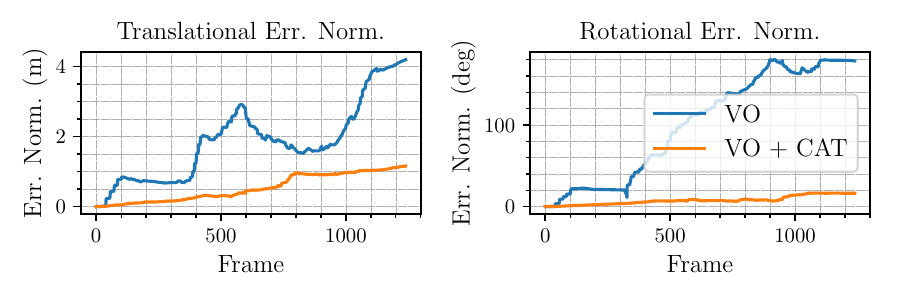}
    \end{subfigure}
    \vspace{-5pt}

    \caption{Comparison of VO errors on the \texttt{ETHL/syn2} trajectory under the ``Global'' (\emph{top row}) and ``Flashlight'' (\emph{bottom row}) conditions. Applying a canonical appearance transformation (CAT) reduces both translational and rotational VO errors (see~\Cref{tab:vo}).}
    \label{fig:vo-errors}
    \vspace{-12pt}
\end{figure}

\Cref{fig:vo-errors} shows sample VO errors for the ``Global'' and ``Flashlight'' conditions of the \texttt{ETHL/syn2} trajectory using both the original image streams (resized and cropped to $256 \times 192$ resolution) and the outputs of our trained CAT model, while \Cref{tab:vo} summarizes our VO results for all \texttt{ETHL} sequences.
Although our VO pipeline successfully tracked the entirety of the \texttt{ETHL} sequences despite rapidly time-varying illumination, the average translational and rotational errors were consistently lower using the CAT models due to the improved temporal consistency of the transformed image stream.
We note that, with the exception of the ``Flashlight'' condition, which differs significantly from the other four, the VO accuracy obtained using trained CAT models is similar to that obtained using the original ``Static'' image stream, indicating that our models produce consistent outputs.

We also investigated the use of a CAT model trained on all \texttt{ETHL/syn} sequences for improving VO on the \texttt{ETHL/real} sequences.
As shown in~\Cref{tab:vo}, the model yielded only small improvements in accuracy compared to the \texttt{ETHL/syn} sequences.
We believe this is because the \texttt{ETHL/real} sequences differ too much from the \texttt{ETHL/syn} sequences in terms of appearance and illumination for the learned model to be useful (see \Cref{fig:ethl_real}).
However, defaulting to a transformation near identity is a desirable property in such cases.

\begin{figure}
    \centering
    
    \begin{subfigure}{0.9\columnwidth}
        \includegraphics[width=\textwidth]{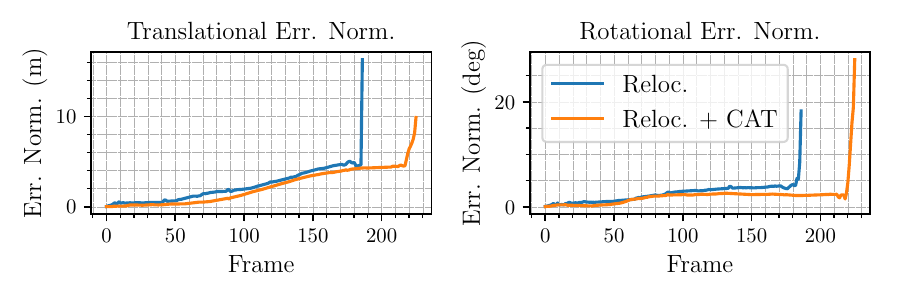}
    \end{subfigure}     
    \vspace{-5pt}
    
    \begin{subfigure}{0.9\columnwidth}
        \includegraphics[width=\textwidth]{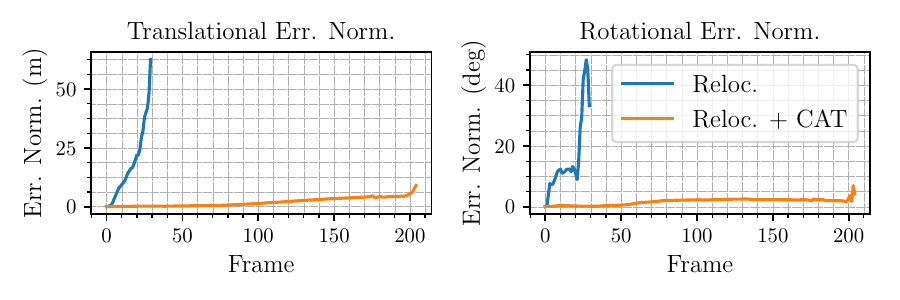}
    \end{subfigure}
    \vspace{-5pt}

    \caption{Comparison of metric relocalization errors on the \texttt{VKITTI/0001} trajectory under the ``Morning'' (\emph{top row}) and ``Sunset'' (\emph{bottom row}) conditions, using a keyframe map created in the ``Overcast'' condition. Applying a canonical appearance transformation (CAT) increases localization success and reduces translational and rotational localization errors (see~\Cref{tab:localization}).}
    \label{fig:reloc-errors}
    \vspace{-12pt}
\end{figure}

\begin{table*}[]
    \centering
    \caption{Comparison of direct visual relocalization against a keyframe map created in the canonical condition, with and without applying a learned canonical appearance transformation (CAT). The best results are highlighted in bold.}
    \label{tab:localization}
    \begin{threeparttable}
    \begin{tabular}{@{}llcccccccc@{}}
        \toprule
        &  & \multicolumn{2}{c}{\textbf{Frames Tracked (\%)}} &  & \multicolumn{2}{c}{\textbf{Avg. Trans. Err. (\% Dist.)}} &  & \multicolumn{2}{c}{\textbf{Avg. Rot. Err. ($\mathbf{\times 10^{-2}}$ deg/m)}} \\ \cmidrule{3-4} \cmidrule{6-7} \cmidrule{9-10}
       \multicolumn{2}{l}{\textbf{Sequence (length)}} & Without CAT & With CAT &  & Without CAT & With CAT &  & Without CAT & With CAT \\ \midrule
       \multicolumn{2}{l}{\texttt{ETHL/syn1} (880 frames, 9.0 m)} &  &  &  &  &  &  &  &  \\
        & Static (canonical) & \textbf{100.00} & \textbf{100.00} &  & \textbf{1.44} & 1.55 &  & 45.17 & \textbf{44.73} \\
        & Local & 98.98 & \textbf{100.00} &  & 1.67 & \textbf{1.55} &  & 48.22 & \textbf{45.28} \\
        & Global & 5.91 & \textbf{100.00} &  & 41.94 & \textbf{1.55} &  & 1151.61 & \textbf{45.17} \\
        & Local + Global & 8.07 & \textbf{100.00} &  & 74.55 & \textbf{1.55} &  & 794.55 & \textbf{45.50} \\
        & Flashlight & 1.59 & \textbf{31.02} &  & 833.33 & \textbf{6.10} &  & 38100.00 & \textbf{97.87} \\ \addlinespace
       \multicolumn{2}{l}{\texttt{ETHL/syn2} (1240 frames, 7.8 m)} &  &  &  &  &  &  &  &  \\
        & Static (canonical) & \textbf{100.00} & \textbf{100.00} &  & \textbf{1.66} & \textbf{1.66} &  & 32.35 & \textbf{31.07} \\
        & Local & 97.02 & \textbf{100.00} &  & 1.58 & \textbf{1.53} &  & \textbf{33.25} & 36.96 \\
        & Global & 4.84 & \textbf{100.00} &  & 206.25 & \textbf{1.53} &  & 3293.75 & \textbf{37.08} \\
        & Local + Global & 5.00 & \textbf{100.00} &  & 223.53 & \textbf{1.53} &  & 3905.88 & \textbf{36.96} \\
        & Flashlight & 1.94 & \textbf{40.08} &  & 887.50 & \textbf{2.51} &  & 38362.50 & \textbf{55.91} \\ \midrule
       \multicolumn{2}{l}{\texttt{VKITTI/0001} (447 frames, 332.5 m)} &  &  &  &  &  &  &  &  \\
        & Overcast (canonical) & \textbf{100.00} & \textbf{100.00} &  & 1.74 & \textbf{1.46} &  & 1.21 & \textbf{0.69} \\
        & Clone & 40.72 & \textbf{100.00} &  & \textbf{1.38} & 1.46 &  & 1.53 & \textbf{0.69} \\
        & Morning & 41.83 & \textbf{50.56} &  & 1.21 & \textbf{1.13} &  & 1.35 & \textbf{1.06} \\
        & Sunset & 6.71 & \textbf{45.86} &  & 56.92 & \textbf{1.04} &  & 52.22 & \textbf{0.91} \\ \addlinespace
       \multicolumn{2}{l}{\texttt{VKITTI/0002} (233 frames, 113.6 m)} &  &  &  &  &  &  &  &  \\
        & Overcast (canonical) & \textbf{100.00} & \textbf{100.00} &  & 11.07 & \textbf{0.92} &  & 4.65 & \textbf{0.97} \\
        & Clone & 13.73 & \textbf{100.00} &  & 43.40 & \textbf{0.92} &  & 32.30 & \textbf{0.94} \\
        & Morning & 46.78 & \textbf{100.00} &  & 25.96 & \textbf{0.93} &  & 11.48 & \textbf{1.01} \\
        & Sunset & 80.26 & \textbf{81.97} &  & 6.58 & \textbf{0.85} &  & 3.11 & \textbf{0.98} \\ \addlinespace
       \multicolumn{2}{l}{\texttt{VKITTI/0006} (270 frames, 51.9 m)} &  &  &  &  &  &  &  &  \\
        & Overcast (canonical) & \textbf{100.00} & \textbf{100.00} &  & 0.69 & \textbf{0.42} &  & 0.44 & \textbf{0.29} \\
        & Clone & 1.48 & \textbf{94.44} &  & 6900.00 & \textbf{1.55} &  & 67800.00 & \textbf{1.16} \\
        & Morning & 84.07 & \textbf{93.70} &  & 2.41 & \textbf{1.52} &  & 6.48 & \textbf{1.91} \\
        & Sunset & 82.59 & \textbf{88.52} &  & 3.32 & \textbf{1.99} &  & \textbf{2.36} & 2.52 \\ \addlinespace
       \multicolumn{2}{l}{\texttt{VKITTI/0018} (339 frames, 254.4 m)} &  &  &  &  &  &  &  &  \\
        & Overcast (canonical) & \textbf{100.00} & \textbf{100.00} &  & \textbf{9.39} & 10.10 &  & \textbf{4.67} & 4.99 \\
        & Clone & 0.59 & \textbf{100.00} &  & 318.34 & \textbf{10.19} &  & 1311.83 & \textbf{5.08} \\
        & Morning & 3.54 & \textbf{67.85} &  & 90.88 & \textbf{6.32} &  & 68.16 & \textbf{3.05} \\
        & Sunset & 10.03 & \textbf{100.00} &  & 11.04 & \textbf{5.29} &  & 21.98 & \textbf{3.54} \\ \addlinespace
       \multicolumn{2}{l}{\texttt{VKITTI/0020} (837 frames, 711.2 m)} &  &  &  &  &  &  &  &  \\
        & Overcast (canonical) & \textbf{100.00} & \textbf{100.00} &  & 8.25 & \textbf{7.87} &  & 1.91 & \textbf{0.90} \\
        & Clone & 4.54 & \textbf{9.20} &  & 189.17 & \textbf{12.75} &  & 22.83 & \textbf{7.93} \\
        & Morning & 19.47 & \textbf{23.89} &  & \textbf{22.25} & 22.50 &  & 2.77 & \textbf{2.71} \\
        & Sunset & 9.44 & \textbf{24.61} &  & 24.81 & \textbf{22.87} &  & 11.54 & \textbf{2.64} \\ \addlinespace
        \multicolumn{2}{l}{\texttt{KITTI/05}\tnote{1}~~(2762 frames, 2206 m)} &  &  &  &  &  &  &  &  \\
        & Clone (map)\tnote{2} & \textbf{100.00} & \textbf{100.00} &  & 1.99 & \textbf{1.65} &  & \textbf{0.38} & 0.80 \\
        & Light & 2.32 & \textbf{18.10} &  & 86.45 & \textbf{2.88} &  & 23.92 & \textbf{1.70} \\
        & Dark & 2.50 & \textbf{5.32} &  & 12.94 & \textbf{1.72} &  & 9.40 & \textbf{5.08} \\ \bottomrule
    \end{tabular}
    \begin{tablenotes}
        \item[1] Model trained on all \texttt{VKITTI} sequences. Data from the \texttt{KITTI/05} sequences were not used for training the models used on the \texttt{VKITTI} sequences
        \item[2] The ``Clone'' condition was used to create the initial keyframe map for relocalization experiments with the \texttt{KITTI/05} sequences.
    \end{tablenotes}
    \end{threeparttable}
\end{table*}

\subsection{Visual relocalization}
In addition to VO, a common task for autonomous vehicles is metric relocalization against an existing map, and an important competency for long-term autonomy is the ability to do so at different times of day when the illumination of the environment may have changed.
To this end, we investigated the use of our trained CAT models for teach-and-repeat-style metric relocalization~\cite{Paton2017-fi,Clement2017-gx} by first creating a map in the canonical condition, then relocalizing against it in different conditions using both the original and transformed images.

\Cref{fig:reloc-errors} shows sample relocalization errors for the ``Morning'' and ``Sunset'' conditions of the \texttt{VKITTI/0001} trajectory using both the original image streams (resized and cropped to $256 \times 192$ resolution) and the output of our trained model, while \Cref{tab:localization} summarizes our relocalization results for all \texttt{ETHL/syn} and \texttt{VKITTI} sequences.
We observed significant improvements in terms of both localization success and localization accuracy using our trained models, with many otherwise untraversable sequences becoming fully or mostly traversable.
In most cases, our method achieved relocalization accuracy on par with that achieved in the VO experiments, suggesting that the output of our models is generally consistent for different inputs.

The outlier in these experiments is the \texttt{VKITTI/0020} sequence, on which our method yielded significantly smaller improvements in localization success rates compared to the other sequences.
This may be due to the fact that much of the camera's field of view in this sequence is occupied by moving vehicles, which makes localization challenging and introduces a significant amount of view-dependent high-frequency reflectance that is not well captured by our model.

We further investigated the use of a CAT model trained on all \texttt{VKITTI} sequences for improving relocalization on the modified KITTI odometry benchmark sequences described in \Cref{sec:vkitti_dataset}.
To do this, we modified \Cref{eq:camera_model,eq:inv_camera_model} to use stereo disparity rather than depth and used a standard block matching algorithm to compute the disparity map.
We used the ``Clone'' condition of the \texttt{KITTI/05} sequence to create the keyframe map for these experiments. 
As shown in \Cref{tab:localization}, the \texttt{VKITTI}-trained model yielded only modest gains in localization success and accuracy compared to the untransformed imagery on sequence \texttt{KITTI/05}.
Returning to \Cref{fig:kitti_affine}, we see that the model outputs are not entirely consistent and contain visible artifacts, which is an indication that the \texttt{VKITTI} training data are not perfectly representative of the real KITTI data.
Further investigation is needed to establish the limits of synthetic-to-real transfer learning in this context, which we leave to future work.

\section{Conclusions} \label{sec:conclusion}
In this work we presented a method for improving the robustness of direct visual localization to environmental illumination change by training a deep convolutional encoder-decoder model to re-illuminate images under a previously-seen canonical appearance.
We validated the use of such canonical appearance transformations (CATs) for both visual odometry (VO) and keyframe-based relocalization tasks using two high-fidelity synthetic \mbox{RGB-D} datasets, and demonstrated significant gains in terms of both localization accuracy and relocalization success under conditions of severe appearance change where conventional methods often fail.

Avenues for future work include investigating how the requirement for identically posed training examples can be relaxed so that training can be easily accomplished outside of simulation, as well as further exploring the use of simulation-trained models in real environments with similar properties, and investigating how networks such as ours can be integrated more deeply into direct visual localization systems.

\bibliographystyle{ieeetr}
\bibliography{refs}

\end{document}